\documentclass[sigconf]{aamas}

\usepackage{balance} 
\usepackage{url}
\PassOptionsToPackage{pagebackref,breaklinks,colorlinks}{hyperref}
\usepackage{amsmath} 
\usepackage{hyperref}
\usepackage{url}
\usepackage{booktabs}
\usepackage[linesnumbered,ruled,vlined]{algorithm2e}

\makeatletter
\gdef\@copyrightpermission{
  \begin{minipage}{0.2\columnwidth}
   \href{https://creativecommons.org/licenses/by/4.0/}{\includegraphics[width=0.90\textwidth]{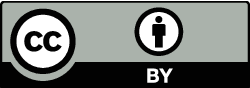}}
  \end{minipage}\hfill
  \begin{minipage}{0.8\columnwidth}
   \href{https://creativecommons.org/licenses/by/4.0/}{This work is licensed under a Creative Commons Attribution International 4.0 License.}
  \end{minipage}
  \vspace{5pt}
}
\makeatother

\setcopyright{ifaamas}
\acmConference[AAMAS '25]{Proc.\@ of the 24th International Conference
on Autonomous Agents and Multiagent Systems (AAMAS 2025)}{May 19 -- 23, 2025}
{Detroit, Michigan, USA}{Y.~Vorobeychik, S.~Das, A.~Nowé  (eds.)}
\copyrightyear{2025}
\acmYear{2025}
\acmDOI{}
\acmPrice{}
\acmISBN{}

\acmSubmissionID{<<OpenReview submission id>>}

\title[AAMAS-2025 Formatting Instructions]{Learning Graph Representation of Agent Diffusers}

\author{Youcef Djenouri}
\affiliation{
  \institution{University of South-Eastern Norway}
  \city{Vestfold}
  \country{Norway}
}
\email{youcef.djenouri@usn.no}
\affiliation{
  \institution{Norwegian Research Centre}
  \city{Oslo}
  \country{Norway}
}
\email{youcef.djenouri@usn.no}

\author{Nassim Belmecheri }
\affiliation{
  \institution{Simula Laboratory Research}
  \city{Oslo}
  \country{Norway}}
\email{nassim@simula.no}

\author{Tomasz Michalak}
\affiliation{
  \institution{University of Warsaw}
  \city{Warsaw}
  \country{Poland}}
\email{tpm@mimuw.edu.pl}

\author{Jan Dubiński}
\affiliation{
  \institution{Warsaw University of Technology}
  \city{Warsaw}
  \country{Poland}}
\email{jan.dubinski.dokt@pw.edu.pl}

\author{Ahmed Nabil Belbachir}
\affiliation{
  \institution{Norwegian Research Centre}
  \city{Grimstad}
  \country{Norway}}
\email{nabe@norcersearch.no}

\author{Anis Yazidi}
\affiliation{
  \institution{University of Oslo}
  \city{Oslo}
  \country{Norway}}
\email{anisy@ifi.uio.no}

\begin{abstract}
Diffusion-based generative models have significantly advanced text-to-image synthesis, demonstrating impressive text comprehension and zero-shot generalization. These models refine images from random noise based on textual prompts, with initial reliance on text input shifting towards enhanced visual fidelity over time. This transition suggests that static model parameters might not optimally address the distinct phases of generation. We introduce LGR-AD (Learning Graph Representation of Agent Diffusers), a novel multi-agent system designed to improve adaptability in dynamic computer vision tasks. LGR-AD models the generation process as a distributed system of interacting agents, each representing an expert sub-model. These agents dynamically adapt to varying conditions and collaborate through a graph neural network that encodes their relationships and performance metrics. Our approach employs a coordination mechanism based on top-k maximum spanning trees, optimizing the generation process. Each agent’s decision-making is guided by a meta-model that minimizes a novel loss function, balancing accuracy and diversity. Theoretical analysis and extensive empirical evaluations show that LGR-AD outperforms traditional diffusion models across various benchmarks, highlighting its potential for scalable and flexible solutions in complex image generation tasks. Code can be found at: \url{https://github.com/YousIA/LGR_AD}.
\end{abstract}


\newcommand{\BibTeX}{\rm B\kern-.05em{\sc i\kern-.025em b}\kern-.08em\TeX}

\begin{document}


\pagestyle{fancy}
\fancyhead{}


\maketitle 

\section{Introduction}
\noindent Diffusion models, or diffusers, have revolutionized the computer vision domain and are the cornerstone of modern text-to-image systems. These models demonstrate impressive capabilities in converting complex textual prompts into photorealistic images, including those involving novel and previously unseen concepts \cite{croitoru2023diffusion,gandikota2023erasing,hudson2024soda}. In addition, diffusers have paved the way for a range of interactive applications, significantly accelerating the democratization of content creation. However, the performance of diffusion models is not consistent across all datasets; certain architectures excel under specific conditions while others underperform.  To address these performance discrepancies, researchers have increasingly explored ensemble methods and expert diffusers, which aggregate the outputs of multiple specialized models to improve results \cite{koley2024text,wang2024ensembling,zeng2024jedi,balaji2022ediff}. Expert diffusers, in particular, have shown superior capabilities by generating high-quality images from both simple and complex text prompts \cite{balaji2022ediff}. Despite these advantages, the adoption of expert diffusers faces two major challenges within the framework of multi-agent systems. The first challenge stems from their resource-intensive nature. Since expert diffusers require loading and executing all constituent models in the expert pool during inference, they impose significant computational demands in terms of time and memory, leading to latency issues. This is especially problematic in multi-agent environments where efficiency and scalability are crucial for real-time applications. The second challenge lies in the collaborative dynamics among models in the expert pool. The inherent interactions between models, analogous to multi-agent coordination, can lead to suboptimal results when certain models exert a counterproductive influence on the collective process. This interference can impede the overall performance gains that expert diffusers are intended to provide, highlighting the need for improved agent communication and coordination strategies to prevent such disruptions.  Addressing these challenges requires further research into optimizing the coordination mechanisms and resource management strategies within multi-agent systems to ensure that the full potential of expert diffusers can be realized in diverse real-world applications.

\paragraph{Motivations}

\textit{We hypothesize that analyzing the diverse interactions among models within the agent diffusers pool can yield valuable insights to enhance the overall text-to-image generation process.} \\

\noindent Our hypothesis suggests that by carefully analyzing and modeling the interactions between agents (i.e., the models in the agent diffusers pool), we can identify the optimal subset of models that positively contribute to the inference process. This approach aligns with the core principles of multi-agent systems, where not all agents contribute equally to a given task, and by selecting the most effective agents, we can significantly improve the system’s overall performance. In this research, we explore the dependencies and synergies among the models using a graph-based multi-agent representation. A graph structure, consisting of vertices and edges, allows us to model the complex relationships between individual models and analyze their interactions. Graph-based representations are powerful tools for capturing interdependencies among entities, enabling us to derive patterns that enhance collective decision-making \cite{peng2023grlc}. In the context of agent diffusers, the goal is to obtain robust representations that offer meaningful insights into how various models collaborate within the system.

\paragraph{Contributions}

This study provides the first comprehensive investigation of graph-based representation in the context of agent diffusers, specifically designed to address the complex coordination challenges posed by modern diffusion models in multi-agent systems. We introduce the LGR-AD framework (Learning Graph Representation of Agent Diffusers), which establishes a novel paradigm for building adaptive diffusion systems that surpass the limitations of both single-task and multi-task models. The key contributions of this paper are as follows:

\begin{enumerate}
    \item We propose a graph-based representation of agent diffusers to capture the diverse features and performance metrics of each model within the pool. To this end, we introduce several strategies for constructing the graph of interacting models, facilitating more efficient selection of the optimal subset of agents.

    \item We develop a \textbf{Graph Convolutional Neural Network (GCNN)} as a meta-model to learn optimal task execution by leveraging the structural properties of the graph. The GCNN minimizes a novel composite loss function that accounts for model diversity, prediction accuracy, and the hierarchical structure of the expert agents. We also provide a theoretical analysis of this loss function, highlighting its effectiveness in multi-agent coordination.

    \item We conduct extensive experiments to evaluate the performance of \textbf{LGR-ED} on well-established benchmarks. Our results demonstrate that LGR-ED consistently outperforms previous state-of-the-art solutions, achieving superior output quality across a range of computer vision tasks.
\end{enumerate}

\section{Related Work}
\paragraph{Diffusion Models} Diffusion models \cite{ho2020denoising,nichol2021improved,yu2023video} represent a significant advancement within the family of generative models. Previously, Generative Adversarial Networks (GANs) \cite{bohavcek2023geometric} were predominantly utilized for generative tasks. However, diffusion and score-based generative models have demonstrated notable improvements, particularly in the domain of image synthesis \cite{ho2022cascaded,zeng2024jedi,cao2024leftrefill}. Several initiatives in the computer vision community attempt to improve the learning of diffusion models by updating the architectures used in image encoding and decoding \cite{chen2023diffusiondet,ji2023ddp,wallace2024diffusion}. Despite the significant success of these models in generating high-quality images across diverse prompt contexts, they may still produce images that do not align with the specified user prompt. 

\paragraph{Model Ensembling and Merging} Model ensembling and merging are effective techniques for enhancing model performance and have been extensively used in various computer vision tasks \cite{zhou2024expandable,tan2024ensemble,al2023unified}. MagicFusion \cite{zhao2023magicfusion} is a notable approach in this domain, focusing on diffusion models by fusing the predicted noises from two expert U-Net denoisers to achieve applications such as style transfer and object binding. Additionally, several intuitive merging-based methods have been developed. A popular approach is Weighted Merging \cite{matena2022merging}, which involves manually assigning weights to merge each U-Net parameter across multiple shared models. Despite its simplicity, Weighted Merging tends to coarsely distribute weights across all U-Net blocks. These ensemble and merging models face limitations due to decoder performance bottlenecks and the extensive time required for enumerative selection to identify optimal settings.

\paragraph{Discussion} In this work, we build on recent advancements in diffusion models by integrating model ensembling and multi-agent coordination strategies. Our approach aims to enhance the capability of diffusion models to generate high-quality images from user prompts by leveraging the collaborative strengths of multiple model ensembles. Specifically, we treat each model in the ensemble as an individual agent within a multi-agent system, where their coordinated interactions improve image fidelity and coherence. To achieve this, we employ advanced merging methods based on graph-based representations to optimally blend the parameters and outputs of these agents. This graph structure captures the interdependencies between the models, enabling more effective communication and coordination. The diffusion process is thus guided by the collective intelligence of the multi-agent system, addressing the limitations of individual models and resulting in superior image generation that meets user requirements with greater precision and detail.

\section{Background}

\paragraph{Text-to-Image Diffusion-Based Generation Problem}

Given a text description \( T \), the goal is to generate a corresponding image \( I \) that visually represents the content of \( T \). This is achieved using a diffusion-based generative model. The diffusion process involves a forward process that progressively adds noise to the image and a reverse process that denoises it to generate a realistic image from random noise, guided by the text description \( T \) \cite{xu2024sgdm,yang2024improving,liu2023instaflow}.

\paragraph{Forward Diffusion Process}
The forward diffusion process is defined as a Markov chain that gradually adds Gaussian noise to the image. Let \( \mathbf{x}_0 \) be the initial image (which is the final output the model aims to learn to generate), and \( \mathbf{x}_t \) be the image at time step \( t \) \cite{cao2024survey,croitoru2023diffusion,yang2023diffusion}. The forward process can be expressed as:$
    q(\mathbf{x}_t | \mathbf{x}_{t-1}) = \mathcal{N}(\mathbf{x}_t; \sqrt{\alpha_t} \mathbf{x}_{t-1}, (1 - \alpha_t) \mathbf{I})$. \( \alpha_t \) is a variance schedule that controls the amount of noise added at each step.

\paragraph{Reverse Diffusion Process}
The reverse diffusion process aims to recover \( \mathbf{x}_0 \) from \( \mathbf{x}_T \) (pure noise) by denoising \cite{gilboa2002forward,chung2023parallel,laroche2024fast}. This process is guided by a model \( p_\theta \) parameterized by \( \theta \), conditioned on the text \( T \):
\begin{equation}
    p_\theta(\mathbf{x}_{t-1} | \mathbf{x}_t, T) = \mathcal{N}(\mathbf{x}_{t-1}; \mu_\theta(\mathbf{x}_t, t, T), \Sigma_\theta(\mathbf{x}_t, t))
\end{equation}
\( \mu_\theta \) and \( \Sigma_\theta \) are the predicted mean and covariance of the Gaussian distribution at step \( t \), conditioned on the noisy image \( \mathbf{x}_t \) and the text \( T \).

\paragraph{Training Objective}
The model is trained to minimize the variational bound on the negative log-likelihood of the data, leading to the following loss function:
\begin{equation}
    L(\theta) = \mathbb{E}_{q} \left[ \sum_{t=1}^{T} D_{\text{KL}}(q(\mathbf{x}_{t-1} | \mathbf{x}_t, \mathbf{x}_0) \| p_\theta(\mathbf{x}_{t-1} | \mathbf{x}_t)) \right]
\end{equation}
where \( D_{\text{KL}} \) denotes the Kullback-Leibler divergence \cite{van2014renyi}.

\paragraph{Generation Process}
To generate an image from a text description \( T \), the process starts with a sample from a Gaussian distribution \( \mathbf{x}_T \sim \mathcal{N}(0, \mathbf{I}) \) and iteratively applies the reverse diffusion steps, 
$\mathbf{x}_{t-1} \sim p_\theta(\mathbf{x}_{t-1}  \mathbf{x}_t, T)
$, until \( \mathbf{x}_0 \) is obtained, which is the final generated image.

\begin{figure*}[ht!]
\centering
\includegraphics[width=\textwidth]{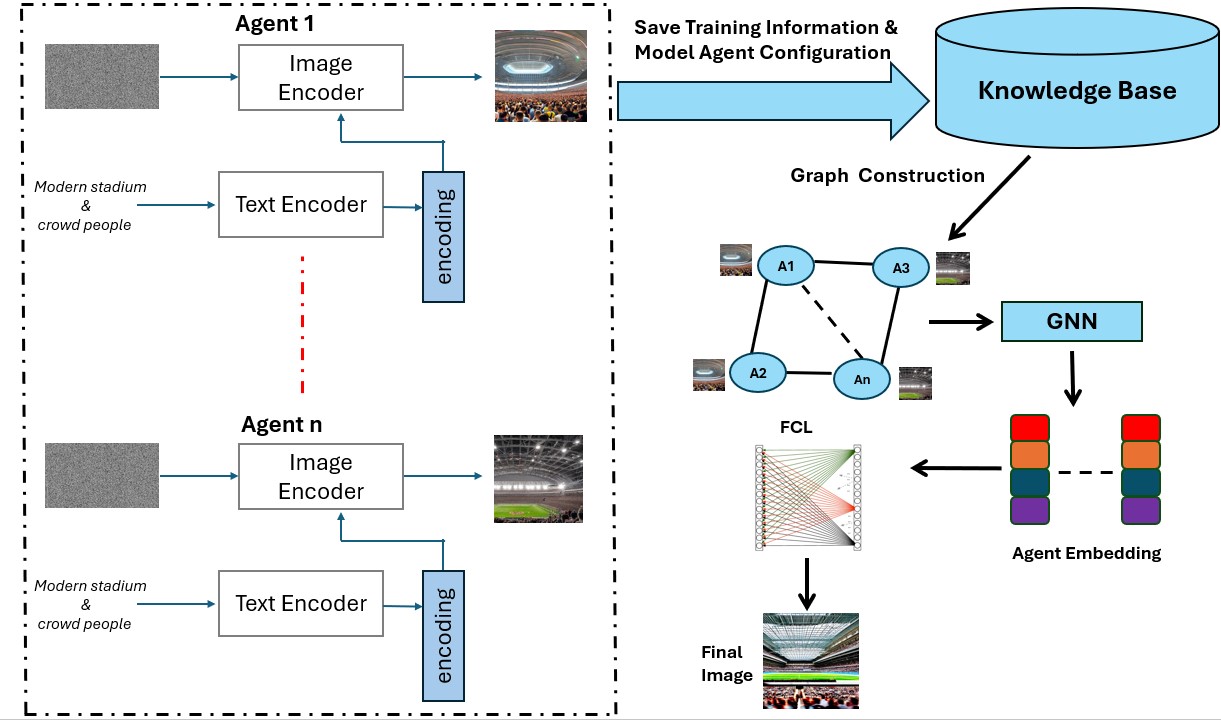}
\caption{\textit{LGR-AD begins by training diffusion-based models for text-to-image generation, treating each model as an agent in a multi-agent system. After training each agent diffuser , we use the agents' outputs and specifications to construct a graph, where nodes represent agents and edges capture their interactions. GCNN is then applied to learn an optimal representation of the graph, leveraging agent collaboration. The learned features are processed through a fully connected layer to guide image generation, enabling the system to adapt and refine the diffusion process for improved fidelity and coherence.}}\label{fig:LGR}
\end{figure*}

\section{LGR-AD: Learning Graph Representation for Agent Diffusers}

\subsection{Principle} 

\noindent In this section, we introduce a novel approach called Learning Graph Representation for Agent Diffusers (LGR-AD), which integrates principles from diffusion models with graph-based representation to enhance the text-to-image generation process. The conceptual framework of LGR-AD is illustrated in Figure \ref{fig:LGR}. This approach comprises several distinct stages, each contributing to the creation of integrated expert diffusers ensemble. Initially, a diverse set of diffusers is trained using appropriate datasets and algorithms, encompassing various architectures and learning strategies. Upon completing the training phase, the resultant outputs and specifications of each individual model are stored within a knowledge base. This repository of model-specific information serves as the foundation for constructing a graph representation. The process of creating the model graph begins with utilizing the accumulated knowledge base. Each trained model is represented as a node in the graph, and relationships between these nodes are established based on observed interconnections and similarities among the models. This results in a graph where nodes denote the trained models, and edges signify the associations between these models, reflecting their shared characteristics and behaviors. Subsequently, a Graph Convolutional Neural Network (GCNN) is deployed to extract insights from the constructed model graph. The primary objective of this step is to capture the correlations and dependencies that exist between the various models. The GCNN transforms the information embedded within the model graph into a lower-dimensional vector space, effectively capturing nuanced relationships that might not be apparent in the original high-dimensional space. The GCNN phase produces an embedded vector that captures the relationships among the trained models. This vector, a condensed representation of the ensemble's collective behavior, is then processed by a fully connected layer to generate a final output image tailored to the user's objective. The subsequent sections provide detailed descriptions of LGR-AD's core components.

\subsection{Graph Representation for Models}
\begin{definition}[Model Output]
We define the set of outputs for model $\mathcal{M}_i$ of the agent $\mathcal{A}_i$ as the union of all its outputs on the training dataset $D$. Formally, we write:
\begin{equation}
  \mathcal{Y}^*_i =\{\bigcup_{D_j \in D} y_{ij}^* \}  
\end{equation}
where, $y_{ij}^*$ is the predicted value of $D_j$ by the model $\mathcal{M}_i$ of the agent $\mathcal{A}_i$. 
\end{definition}

\begin{definition}[Model Specification]
We define the set of model specification $\mathcal{S}_i$ of the model $\mathcal{M}_i$ by the set of the representative layers of the model $\mathcal{M}_i$. For instance, $\mathcal{S}_i=\{conv = 1, pool=1, att=0, bn=0, dr=1\}$ to represent that the model $\mathcal{M}_i$ has a convolution layer, a pooling layer, no attention layer, no batch normalization layer, and has a dropout layer.
\end{definition}

\begin{definition}[Connectivity Function]
Consider a set of $n$ models $\mathcal{M}=\{\mathcal{M}_1, \mathcal{M}_2...\mathcal{M}_n\}$, each model $\mathcal{M}_i$ represents the behaviour of the agent $\mathcal{A}_i$, we define $\displaystyle f_{ij}$  a function that connects the two models $\mathcal{M}_i$, and $\mathcal{M}_j$, and we write: 
$$f: \mathcal{M} \times \mathcal{M} \rightarrow \mathbb{R}$$
\end{definition}

\begin{definition}[Characteristic Connectivity Function ]
We define the Characteristic Connectivity Function (CCF) as a connectivity function that compute the similarity between two models specification, and we write: 
\begin{equation}
    CCF(\mathcal{M}_i, \mathcal{M}_j)= 
    |S_{i} \cap S_{j}| 
\end{equation}
\end{definition}

\begin{definition}[Performance Connectivity Function]
We define the Performance Connectivity Function (PCF) as a connectivity function that computes the similarity between two models output, and we write: 

\begin{equation}
    PCF(\mathcal{M}_i, \mathcal{M}_j)= |Y_{i}^* \cap Y_{j}^*| 
\end{equation}
\end{definition}

\begin{definition}[Graph of Models]
We define the graph of models $\mathcal{G}_M= (V,E)$ by the set of nodes $V$ represented the set of models $\mathcal{M}$, and a set of edges, $E \subseteq V \times V$. $(\mathcal{M}_i, \mathcal{M}_j )\in E$, if and only if,  $f_{ij} \neq 0$. $f_{ij}$ is calculated as $CCF(\mathcal{M}_i, \mathcal{M}_j)$ or $PCF(\mathcal{M}_i, \mathcal{M}_j)$ depending to the connectivity function chosen by the user. 
\end{definition}

 \subsection{Maximum Spanning Tree for Agent Diffusers}

Let \( G = (\mathcal{A}, E) \) be a graph, where \( \mathcal{A} \) is the set of agents and \( E \) represents the edges (relationships) between these agents. Each edge \( e_{ij} \in E \) between agents \(\mathcal{A}_i \) and \( \mathcal{A}_j \) is weighted by a similarity function \( w(e_{ij}) \) that quantifies the shared properties between the two agents, based on their performance metrics or model characteristics. The objective of the Maximum Spanning Tree (MST) algorithm is to select a subset of edges \( E' \subseteq E \) such that, \( T = (\mathcal{A}, E') \) is a tree (i.e., an acyclic, connected graph), and the total weight of the tree is maximized:  
\begin{equation}
    \max_T \sum_{e_{ij} \in E'} w(e_{ij})  
\end{equation}

The following formal description captures the distinct roles of the MST in LGR-AD:

\noindent \textbf{1. Relationships Capture}: The MST emphasizes the most significant relationships among agents, capturing the most relevant shared characteristics. For each pair of agents \( \mathcal{A}_i, \mathcal{A}_j \in \mathcal{A} \), we define a weight function \( w(e_{ij}) \) that assigns a scalar value based on the similarity of their outputs. The MST maximizes the sum of these weights:

\begin{equation}
T^* = \arg\max_T \sum_{e_{ij} \in E'} w(e_{ij})
\end {equation}

where \( T^* \) is the optimal tree structure that spans all agents with the most significant relationships. By focusing on the highest weights, the MST filters out weaker or less relevant connections, providing a concise representation of the most critical relationships.

\noindent \textbf{2. Reduced Redundancy:}  The MST algorithm prevents cycles, ensuring that redundant connections between agents are removed. This is achieved by enforcing the acyclic property of the tree. Let \( T = (\mathcal{A}, E') \) represent the maximum spanning tree, where \( E' \subseteq E \) is the set of selected edges. The acyclic property ensures:
\[
\forall \, \mathcal{A}_i, \mathcal{A}_j \in T, \, \text{there exists at most one path from } \mathcal{A}_i \text{ to } \mathcal{A}_j.
\]
This property reduces redundancy, capturing only the most informative relationships among agents.

\noindent \textbf{3. Hierarchical Information:}   The hierarchical structure of the MST reflects the importance of relationships between agents. The tree defines a hierarchy of shared properties, starting with general properties and refining down to more specific ones. For any pair of agents \( \mathcal{A}_i \) and \( \mathcal{A}_j \), their depth in the tree reflects the degree of shared characteristics. Formally, the hierarchy can be captured by defining the depth \( d(\mathcal{A}_i) \) of each node \( \mathcal{A}_i \in \mathcal{A} \) in the tree:

\begin{equation}
d(\mathcal{A}_i) = \min_{\text{paths}} \sum_{e_{ij} \in T} w(e_{ij})
\end{equation}

where deeper nodes in the tree represent agents with more specialized relationships. The hierarchical nature of the MST provides insights into how closely the agents are related, both in general and in specific terms.

\noindent \textbf{4. Ensemble Composition:}   
By selecting only the most significant edges in the MST, we form a focused and diverse ensemble of models. The ensemble \( \mathcal{E} \) consists of agents connected by the edges of the MST. This selective process can be formulated as:

\begin{equation}
\mathcal{E} = \{\mathcal{A}_i \mid \mathcal{A}_i \in T^*\}
\end{equation}

The collective performance of this ensemble is enhanced by emphasizing relationships that provide complementary strengths across agents. By maximizing the relevant shared properties, the ensemble is designed to leverage the diversity of the agents, leading to improved performance across a variety of tasks.

\subsection{Intuition of the loss function in Agent Diffusers}
\noindent An important contribution of the LGR-AD algorithm is the introduction of a novel loss function, \( L(x) \) used to train Graph Convolutional Neural Network. 

\begin{definition}[Loss Function]
We define \( L(x) \)  as:
\begin{equation}
    L(x) = C(x) + \lambda D(x) + \gamma L_{\text{laplace}}
\end{equation}
where \(x \) is the input image text, \( C(x) \) is the Cross-Entropy loss, \( D(x) \) is the Kullback-Leibler Divergence loss, \( \lambda \) and \( \gamma \) are regularization parameters. The \( L_{\text{laplace}} \) is a modified version of the Laplacian loss to ensure the model's predictions align with the underlying graph structure \cite{lai2017deep}.
\end{definition}

\noindent To promote diversity, the Kullback-Leibler (KL) Divergence loss needs to be maximized. To incorporate this into the overall loss function, which requires minimization, we introduce the term \(D(x)\). This adjustment allows us to integrate the KL Divergence effectively into the global loss function. The resulting loss function combines the benefits of accuracy, diversity, and structural alignment, providing a comprehensive and versatile optimization objective.

\paragraph{Diversity and Accuracy} The Cross-Entropy loss, calculated between the final generated image and the ground truth image, ensures that the agent of diffusers is accurate, while the Kullback-Leibler Divergence enforces diversity among the models.
\paragraph{Regularization} The \( \lambda \) term acts as a regularization parameter. A higher \( \lambda \) gives more importance to the Kullback-Leibler Divergence, encouraging the ensemble to pay more attention to the diversity of selected agent diffusers.

\paragraph{Laplacian Loss} The Laplacian loss, derived from the Laplacian matrix of the graph, encourages the model to respect the graph's structure. In the context of Maximum Spanning Tree, this loss ensures that the model's predictions align with the underlying tree structure, promoting a more structured prediction.
\paragraph{Trade-off} The hybrid loss allows for a balance between text-to-image generation accuracy, distributional similarity, and graph structure adherence, offering the 'best of both worlds'.
\paragraph{Robustness} The hybrid loss can make the ensemble more robust. While Cross-Entropy loss ensures text-to-image generation accuracy, the Kullback-Leibler part can make the ensemble robust to variations in the input distribution.
\paragraph{Optimal Weighting} The hybrid loss can provide a good performance metric, allowing for potentially more effective weighting of individual models in the ensemble.

\subsection{Graph Convolutional Neural Network (GCNN) for Agent Diffusers}

In this section, we propose the use of a GCNN to effectively model the relationships between agents in the LGR-AD framework. MST, which captures the most significant relationships among the agents, is utilized as the underlying structure of the graph. The GCNN learns a representation of each agent's state, allowing the agents to communicate and collaborate, thus improving their collective performance. The weights of the graph $G$ on the edges are derived from the MST, with each edge weight \( w(e_{ij}) \) representing the strength of the relationship between agents \( v_i \) and \( v_j \). GCNN layer updates the feature vector of each agent by aggregating information from its neighbors in the graph. Formally, the update rule for each GCNN layer is given by:

\begin{equation}
\mathbf{H}^{(l+1)} = \sigma\left( \mathbf{\tilde{M}} \mathbf{H}^{(l)} \mathbf{W}^{(l)} \right)
\end{equation}

where $M$ is the adjacency matrix between the different agents. \( \mathbf{H}^{(l)} \in \mathbb{R}^{|V| \times d^{(l)}} \) is the feature matrix at layer \( l \), with \( d^{(l)} \) being the dimensionality of the feature space at layer \( l \). \( \mathbf{\tilde{M}} = \mathbf{D}^{-\frac{1}{2}} \mathbf{M} \mathbf{D}^{-\frac{1}{2}} \) is the normalized adjacency matrix, where \( \mathbf{D} \) is the diagonal degree matrix with \( \mathbf{D}_{ii} = \sum_j \mathbf{M}_{ij} \). \( \mathbf{W}^{(l)} \in \mathbb{R}^{d^{(l)} \times d^{(l+1)}} \) is the learnable weight matrix at layer \( l \). \( \sigma(\cdot) \) is a non-linear activation function, such as ReLU. 

The GCNN processes multiple layers of graph convolutions, allowing each agent to aggregate information from its local neighborhood in the graph. After \( L \) layers, the output matrix \( \mathbf{H}^{(L)} \in \mathbb{R}^{|+mathcal{A}| \times d^{(L)}} \) represents the learned embeddings for each agent, capturing both their individual properties and the relationships with other agents as encoded by the MST. For the final prediction, a fully connected layer is applied to each node's embedding to generate the output for each agent:

\begin{equation}
\mathbf{y}_i = \text{softmax}\left( \mathbf{W}^{\text{out}} \mathbf{h}_i^{(L)} + \mathbf{b}^{\text{out}} \right)
\end{equation}

where \( \mathbf{W}^{\text{out}} \in \mathbb{R}^{d^{(L)} \times c} \) is the weight matrix for the final classification or regression task, \( \mathbf{b}^{\text{out}} \) is the bias term, and \( c \) is the number of output classes or the dimensionality of the regression target.

MST provides the structural backbone for the GCNN, ensuring that the most relevant and informative relationships among agents are emphasized. By using the MST-based adjacency matrix, the GCNN learns to propagate information primarily along the most critical edges, leading to a more efficient and focused learning process. This hierarchical and non-redundant structure enables the system to leverage the complementary strengths of the agents, improving the overall performance of the model.

\subsection{Pseudo-Code}

\begin{algorithm}
\scriptsize 
\DontPrintSemicolon
\KwData{Set of diffusers \( \mathcal{M} \), Training data \( \mathcal{D} \)}
\KwResult{Trained LGR-AD model}
\BlankLine

\BlankLine

\BlankLine

Initialize GCNN model with parameters \( \theta \)\;
\BlankLine

\BlankLine

\For{each epoch}{
    \For{each \( x \in \mathcal{D} \)}{
    
\( G \gets \emptyset \) \tcp*[r]{Initialize the graph}

\ForEach{\( (m_i, m_j) \in \mathcal{M}^2 \) s.t \( i < j \)}{
    \( w_{ij} = f(m_i(x), m_j(x)) \) \tcp*[r]{Compute weight between models using \( f \)}
    \( G \gets G \cup \{(m_i, m_j, w_{ij})\} \) \tcp*[r]{Add edge with weight to the graph}
}
        \( \text{MST} \leftarrow \) Extract $k$ Maximum Spanning Trees from \( G \)
    \BlankLine
         \( A_{MST},\mathbf{h} \gets GCNN(MST) \) 
        
        \( L_{\text{laplace}} \gets \frac{1}{2} \sum_{i,j} \mathbf{A}_{\text{MST}_{ij}} \|\mathbf{h}_i - \mathbf{h}_j\|^2 \) 

        \( L(x) = C(x) + \lambda D(x) + \gamma L_{\text{laplace}} \)
        Compute the loss \( L(x) \)\;
        Compute gradients \( \nabla L(\theta) \)\;
        Update model parameters \( \theta_{t+1} = \theta_t - \eta \nabla L(\theta_t) \)\;
    }
}
\BlankLine
\Return Trained GCNN model with parameters \( \theta \)\;

\caption{LGR-AD Algorithm}
\end{algorithm}
The LGR-AD algorithm enhances the performance of a set of "agent diffusers" for text-to-image generation by explicitly modeling the relationships between them. It begins by constructing a graph where each node represents a diffuser from the set (\( \mathcal{M} \)).  The edges between these nodes are weighted based on a function \( f \) that assesses the relationship or similarity between the outputs of the two connected diffusers (using training data \( \mathcal{D} \)).  To focus on the most important connections, Maximum Spanning Tree (MST) is extracted from this graph, the number of MSTs can be too many, and thus increasing the computational cost since one  MST is built in $O(m.log \ n)$, when all of the MSTs are considered it will be $O(d.m.log \ n)$ where $d$ is the all the number of trees and $m$ is the number of edges and $n$ is the number of vertices. To overcome this only a small number $k$ of MSTs is sampled thus resulting in $O(k.m.log n)$ where $k$ is a bounded number of MSTs (in our case it is 1) .  
A Graph Convolutional Neural Network (GCNN) is then employed to learn from this MST. LGR-AD introduces a "Laplacian loss" ($L_{\text{laplace}} = \frac{1}{2} \sum_{i,j} \mathbf{A}_{\text{MST}_{ij}} \|\mathbf{h}_i - \mathbf{h}_j\|^2$) that encourages the GCNN to learn similar embeddings (\(\mathbf{h}_i\)) for diffusers that are strongly connected in the MST. Here,  \(\mathbf{A}_{\text{MST}_{ij}}\)  is an element in the adjacency matrix of the MST, indicating the connection strength. This loss, in essence, promotes smoothness in the embedding space, ensuring that closely related diffusers have similar representations.

The GCNN is trained using a hybrid loss function that combines this Laplacian loss with the primary task loss (e.g., image reconstruction error) and a regularization term. This combined loss ensures that the model achieves good performance on the main task while also respecting the relationships between the diffusers captured in the MST. By leveraging these relationships, LGR-AD facilitates a more focused and effective learning process, leading to improved performance in text-to-image generation.

\section{Experimental Studies}

\subsection{Experimental Settings}

\noindent In this study, we employed an ensemble of agent diffusion models to achieve high-quality image synthesis. Each model in the ensemble was carefully selected to cover a diverse range of capabilities and specialties. Below, we outline the configuration and hyperparameters for each of the expert models used in our ensemble.

\paragraph{Model Configurations} Table \ref{tab:models} provides an overview of the expert models used in our ensemble, including their model IDs, the number of parameters, and the datasets on which they were pre-trained.

\begin{table}[ht!]
\centering
\scriptsize
\caption{Configurations of Agent Diffusers. All datasets are large image-text pairs collections.}
\label{tab:models}
\begin{tabular}{@{}lcc@{}}
\toprule
\textbf{Model} & \textbf{Param.(B)} & \textbf{Pre-training Dataset} \\ \midrule
DALL-E 2 \cite{ramesh2022hierarchical} & 12 & Internal OpenAI \\
Stable Diffusion v2 \cite{rombach2022high} & 1.5 & Filtered subset of LAION-5B \cite{laion5b}\\
LDM \cite{rombach2022high} & 0.4 & LAION-400M \cite{laion400m} \\
Imagen \cite{saharia2022photorealistic} & 1 & COYO-700M \cite{coyo700m}; internal Google \\ \bottomrule
\end{tabular}
\end{table}

\paragraph{Inference Settings} Table \ref{tab:settings} summarizes the inference settings, including the number of inference steps and the guidance scale used during image generation.

\begin{table}[ht!]
\centering
\scriptsize
\caption{Inference Settings}
\label{tab:settings}
\begin{tabular}{@{}lcc@{}}
\toprule
 & \multicolumn{2}{c}{\textbf{Inference Settings}} \\ \cmidrule(lr){2-3}
\textbf{Model} & \textbf{Num Inference Steps} & \textbf{Guidance Scale}   \\ \midrule
DALL-E 2 \cite{ramesh2022hierarchical} & 100 & 10.0 \\
Stable Diffusion v2 \cite{rombach2022high}  & 50 & 7.5\\
LDM \cite{rombach2022high}  & 75 & 9.0 \\
Imagen \cite{saharia2022photorealistic} & 60 & 8.5  \\ 
\bottomrule
\end{tabular}
\end{table}

\noindent The performance of the ensemble was evaluated based on well-known metrics including Fréchet Inception Distance (FID) \cite{jung2021internalized}, Inception Score (IS) \cite{chong2020effectively}, and CLIP score \cite{zhang2023text}. 

\noindent We evaluated our approach using a suite of datasets that are widely recognized and frequently utilized in the literature \cite{xu2018attngan,zhang2021cross,zhu2019dm}. The selected datasets encompass a diverse range of images and include (MSCOCO (Microsoft Common Objects in Context) \cite{cho2014learning}, CUB (Caltech-UCSD Birds-200-2011) \cite{wah2011caltech}, LN-COCO (Large-scale Noisy-COCO) \cite{pont2020connecting}, Multi-modal CelebA-HQ (MM CelebA-HQ) \cite{xia2021tedigan}). A detailed description of the datasets can be found in the Appendix.

\subsection{Numerical Results}

\begin{table}[ht!]
\centering
\scriptsize
\caption{Comparison of the SOTA text-to-image generation models on MSCOCO.}
\begin{tabular}{cccc}
\toprule
Models & FID $\downarrow$ & IS 	$\uparrow$ & CLIP Score $\uparrow$\\
\midrule
StableDiffusion \cite{rombach2022high} & 9.91 & 85& 0.31 \\
LatteGAN \cite{lattegan} & 11.05 & 74 & 0.28 \\
DALL-E 2 \cite{ramesh2022hierarchical} & 9.91 & 112 & 0.30  \\
Stable Diffusion v2 \cite{rombach2022high}  & 9.78 & 89 & 0.33 \\
LDM \cite{rombach2022high} & 10.32 & 83 & 0.28 \\
Imagen \cite{saharia2022photorealistic} & 9.95 & 116 & 0.35 \\
MagicFusion \cite{zhao2023magicfusion}& 10.94 & 92& 0.29 \\
\textbf{LGR-AD (Our)} & \textbf{9.52} & \textbf{129} & \textbf{0.36} \\ 
\bottomrule
\end{tabular}
\label{tab:ms_coco}
\end{table}

\begin{table}[ht!]
\centering
\scriptsize
\caption{Comparison of the SOTA text-to-image generation models on CUB.}
\begin{tabular}{cccc}
\toprule
Models & FID $\downarrow$ & IS 	$\uparrow$ & CLIP Score $\uparrow$ \\
\midrule
StableDiffusion \cite{rombach2022high} & 10.15 & 82 & 0.39\\
LatteGAN \cite{lattegan} & 10.89  & 75 & 0.42\\
DALL-E 2 & 9.95 & 79 & 0.44 \\
Stable Diffusion v2 \cite{rombach2022high}  & 9.74 &  84 & 0.41\\
LDM \cite{rombach2022high} & 10.33 & 79  & 0.34\\
Imagen \cite{saharia2022photorealistic}  & 10.15 & 93  & 0.44\\
MagicFusion \cite{zhao2023magicfusion}& 9.91  & 77 & 0.40\\
\textbf{LGR-AD (Our)} & \textbf{9.55} & \textbf{105} & \textbf{0.48}\\ 
\bottomrule
\end{tabular}
\label{tab:CUB}
\end{table}

\begin{table}[ht!]
\centering
\scriptsize
\caption{Comparison of the SOTA text-to-image generation models on LN-COCO.}
\begin{tabular}{ccccc}
\toprule
Models & FID $\downarrow$ & IS 	$\uparrow$ & CLIP Score $\uparrow$ \\
\midrule
StableDiffusion \cite{rombach2022high} & 9.66 & 115 & 0.39 \\
LatteGAN \cite{lattegan}& 10.85 & 109 & 0.37   \\
DALL-E 2 \cite{ramesh2022hierarchical} & 9.87 & 118 & 0.40 \\
Stable Diffusion v2 \cite{rombach2022high} & 9.50 & 123 & 0.42  \\
LDM \cite{rombach2022high} & 10.25 & 107 & 0.36 \\
Imagen \cite{saharia2022photorealistic} & 9.89 & 135 &  0.43 \\
MagicFusion \cite{zhao2023magicfusion}& 10.71 & 111 & 0.38 \\
\textbf{LGR-AD (Our)} & \textbf{9.44} & \textbf{149} &\textbf{0.45}\\ 
\bottomrule
\end{tabular}
\label{tab:ln_coco}
\end{table}

\begin{table}[ht!]
\centering
\scriptsize
\caption{Comparison of the SOTA text-to-image generation models on MM CelebA-HQ.}
\begin{tabular}{cccc}
\toprule
Models & FID $\downarrow$ & IS 	$\uparrow$ & CLIP Score $\uparrow$ \\
\midrule
StableDiffusion \cite{rombach2022high} & 9.70 & 154 & 0.49\\
LatteGAN \cite{lattegan} & 10.13 & 141 & 0.48\\
DALL-E 2 \cite{ramesh2022hierarchical} & 9.91 & 144 & 0.54\\
Stable Diffusion v2 \cite{rombach2022high} & 9.57 & 172 & 0.51\\
LDM \cite{rombach2022high} & 10.15& 123  & 0.51\\
Imagen \cite{saharia2022photorealistic} & 9.96 & 185 & 0.53\\
MagicFusion \cite{zhao2023magicfusion}& 9.83 & 129 &  0.52\\
\textbf{LGR-AD (Our)} & \textbf{9.50} & \textbf{190}  & \textbf{0.57}  \\ 
\bottomrule
\end{tabular}
\label{tab:mm}
\end{table}

\begin{table}[ht!]
\centering
\scriptsize
\caption{Comparison of Methods on Diversity Metric}
\begin{tabular}{lllc}
\hline
\textbf{Method}        & \textbf{Backbone}      & \textbf{Dataset}    & \textbf{Diversity↑} \\ \hline
InstantBooth   \cite{shi2024instantbooth}        & Stable Diffusion       & PPR10K              & 0.45                \\ 
DM-GAN  \cite{zhu2019dm}               & ResNet-101            & CUB-200             & 0.55                \\ 
LAFITE  \cite{yin2023lafite}               & ViT-B                 & FashionGen          & 0.58                \\ 
Stable Diffusion  \cite{rombach2022high}     & ViT-L                 & FashionGen          & 0.65                \\ 
VQGAN-CLIP  \cite{crowson2022vqgan}         & ViT-B                 & Oxford Pets         & 0.52                \\ 
T2I-Adapter  \cite{mou2024t2i}          & ViT-B                 & Emogen              & 0.63                \\ 
\textbf{Ours}          & \textbf{ViT-L}        & \textbf{COCO}       & \textbf{0.67}       \\ \hline
\end{tabular}
\label{tab:diversity_results}
\end{table}

\begin{figure*}[ht!]
\centering
\includegraphics[width=16cm]{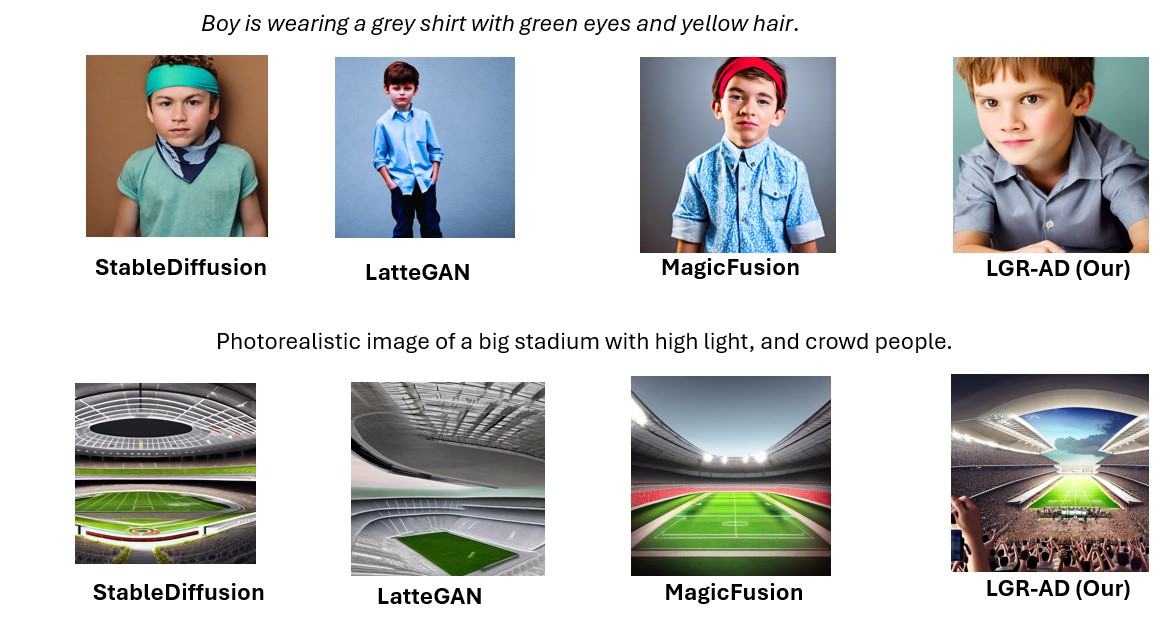}
\caption{\textit{Selected qualitative results of LGR-AD compared to the baseline solutions.}} \label{fig:quality}
\end{figure*}

\noindent We compare the image generation capabilities of our model against existing baselines, including both stand-alone models and state-of-the-art model ensemble method. The results are presented in Tables \ref{tab:ms_coco}, \ref{tab:CUB}, \ref{tab:ln_coco}, and \ref{tab:mm}. Our findings consistently show that LGR-AD achieves superior performance across all datasets and evaluation metrics. Lower FID scores indicate that LGR-AD generates images with a distribution closer to the real images, suggesting high quality and realism. Higher IS values reflect better diversity and perceptual quality of the generated images.  Finally, higher CLIP scores indicate a better alignment between the generated images and their corresponding text descriptions, demonstrating the model's effectiveness in capturing semantic content. These results underscore the usefulness of graph-based representations in diffusion models ensembles. The graph-based approach employed by LGR-AD allows for structured and context-aware representations, leading to an enhanced image quality and a better alignment with textual descriptions. Our proposed methodological innovation not only improves image generation performance on standard benchmarks but also highlights the potential for broader applicability in various computer vision tasks. The superior performance of LGR-AD across multiple datasets and evaluation metrics establishes it as a robust and versatile model for image generation, setting a new standard for future research in this domain. Table \ref{tab:diversity_results} compares several text-to-image generation methods based on the diversity metric across various datasets and backbone architectures. The diversity metric is defined as the average pairwise distance between generated image embeddings, with higher values indicating greater diversity among generated samples. The datasets range from domain-specific collections like CUB-200 (birds) and FashionGen (fashion) to general-purpose datasets like COCO, offering a broad evaluation scope. Ours achieves the highest diversity score of $0.67$ using the ViT-L backbone and COCO dataset, indicating the method`s ability to produce a wide variety of images while preserving coherence to the input text. Classical methods such as DM-GAN, which employs a ResNet-101 backbone, yield relatively lower diversity ($0.55$) due to limitations in capturing nuanced text-to-image mappings. This achievement stems from the seamless integration of multi-expert collaboration and dynamic coordination mechanisms facilitated by graph neural networks.

\subsection{Qualitative Results}
\noindent As depicted in Figure \ref{fig:quality}, we conduct a qualitative comparison between the outputs of the baseline methods, and our proposed LGR-AD approach. This analysis yields several significant observations. Firstly, LGR-AD exhibits a remarkable ability to generate images with enhanced aesthetic quality. Specifically, LGR-AD consistently produces images with superior composition and finer details compared to established methods such as StableDiffusion \cite{rombach2022high}, LatteGAN \cite{lattegan}, and MagicFusion \cite{zhao2023magicfusion}. Secondly, LGR-AD demonstrates a significant advantage in achieving superior context alignment. For instance, all baseline methods generate images depicting an empty stadium, failing to capture the context accurately. In contrast, our LGR-AD method uniquely generates the image of a stadium filled with a crowd, accurately reflecting the provided context and thereby demonstrating its superior contextual understanding. These superior results are attributed to the advanced learning representation mechanisms employed by LGR-AD. The model effectively leverages the correlations among various diffusers, which enhances the learning process and leads to a significant improvement in the quality of the generated images. This efficient representation learning enables LGR-AD to outperform existing methods in both aesthetic quality and contextual alignment, as evidenced by the comparative analysis.

\subsection{Ablation Study}

\begin{table}[ht!]
\centering
\scriptsize
\caption{Varying the Number of Models of LGR-AD.}
\begin{tabular}{lccc}
\toprule
Models & FID $\downarrow$ & IS 	$\uparrow$ & CLIP Score $\uparrow$\\
\midrule
(DALL-E 2, Stable Diff. v2) & 10.38 & 119 & 0.38 \\
(DALL-E 2, LDM) & 10.51 & 115 & 0.36 \\
(DALL-E 2, Imagen) & 9.91 & 125 &  0.38 \\
(Stable Diff. v2, LDM) & 10.31 & 121 & 0.35 \\
(Stable Diffusion v2, Imagen) & 9.95 & 131 &  0.39\\
(Imagen, LDM) & 9.96  & 128 & 0.35 \\
(DALL-E 2, Stable Diff. v2, Imagen) & 9.69 & 139 &  0.43\\
(DALL-E 2, Stable Diff. v2, LDM) & 9.75 & 130 &  0.40\\
(DALL-E 2, Imagen, LDM) & 9.80  & 129 &  0.40\\
(Stable Diff. v2, Imagen, LDM) & 9.81 & 133 &  0.42\\
\textbf{LGR-AD (All models)} & \textbf{9.50} & \textbf{143} &  \textbf{0.46}\\ 
\bottomrule
\end{tabular}
\label{tab:ablation_1}
\end{table}

\begin{table}[ht!]
\centering
\scriptsize
\caption{Varying the Connectivity Function of LGR-AD.}
\begin{tabular}{cccc}
\toprule
Models & FID $\downarrow$ & IS 	$\uparrow$ & CLIP Score $\uparrow$\\
\midrule
CCF & 9.55 & 139 &  0.44 \\
PCF &  9.53 & 140 & 0.44 \\
Hybrid & \textbf{9.50} & \textbf{143} &  \textbf{0.46}\\
\bottomrule
\end{tabular}
\label{tab:ablation_2}
\end{table}

\noindent In our ablation study, we investigate various configurations of LGR-AD to identify the optimal setup for high-quality image generation. Specifically, we experiment with the number of models used in the expert diffusers, ranging from two to four models, and examined different connectivity functions utilized in the graph construction. The models included in the four-model configuration were Stable Diffusion v2, Imagen, DALL-E 2, and LDM. The connectivity functions used are: CCF, PCF, and a Hybrid function that combines both. The results presented in Table \ref{tab:ablation_1} demonstrate a substantial advantage in utilizing all four models (Stable Diffusion v2, Imagen, DALL-E 2, and LDM) compared to using only a subset of these models. This is explained by the fact that each model contributes unique strengths, and their combined use results in a more robust generation process. The results shown in Table \ref{tab:ablation_2} also highlight the importance of using both connectivity functions (CCF and PCF) to better capture the various correlations and dependencies among the agent diffusers in LGR-ED. Indeed, incorporating both CCF and PCF provides a more nuanced understanding of the relationships between different agent diffusers, leading to superior image generation performance. These findings highlight the importance of exploring diverse model architectures and advanced connectivity functions in LGR-AD, pushing the boundaries of image generation quality in computer vision.

\section{Conclusion}
\noindent In this paper, we introduced, LGR-AD, a novel multi-agent system that integrates diffusion-based generative models with graph representation techniques for text-to-image generation. Our approach models the generation process as a distributed system of interacting agents, where each agent represents an expert sub-model specializing in different aspects of the task. The outputs of these agents are stored in a knowledge base, which is then used to construct a graph that encodes the relationships and dependencies among the agents. GCNN learns the complex interactions between agents from this graph, while a fully connected layer synthesizes the final image based on the user's input prompt. Each agent operates with a degree of autonomy, and their collective decision-making is optimized through a meta-model that minimizes a novel loss function. This loss function balances diversity and accuracy across the agents and incorporates a maximum spanning tree approach to enhance coordinated optimization. Our theoretical analysis sheds light on how the graph structure and loss function promote effective collaboration between agents, leading to emergent behaviors that improve overall system performance. Empirical results demonstrate that LGR-AD outperforms traditional diffusion models across various computer vision benchmarks by leveraging multiple specialized agents. This adaptable and scalable multi-agent framework addresses complex tasks effectively. Future work will focus on enabling agents to specialize in subtasks while sharing knowledge, enhancing robustness and adaptability. Additionally, incorporating game-theoretic strategies for agent collaboration could further improve contextual text-to-image generation, highlighting LGR-AD’s potential to advance adaptive image synthesis.

\begin{acks}
This work is funded by the Research Council of Norway under the project entitled "Next Generation 3D Machine Vision with Embedded Visual Computing" under grant number 325748. It is also partially funded by the European Commission through the AI4CCAM project (Trustworthy AI for Connected, Cooperative Automated Mobility) under grant agreement number 101076911, and the National Science Centre in Poland, under grant agreement number 2020/39/O/ST6/01478.
\end{acks}



\bibliographystyle{ACM-Reference-Format} 
\bibliography{main}

\end{document}